\documentclass{article}

\usepackage[nonatbib,preprint]{neurips_2023}

\usepackage[utf8]{inputenc} 
\usepackage[T1]{fontenc}    
\usepackage{hyperref}       
\usepackage{url}            
\usepackage{booktabs}       
\usepackage{amsfonts}       
\usepackage{nicefrac}       
\usepackage{microtype}      
\usepackage{xcolor}         

\usepackage{multirow}
\usepackage{graphicx}
\usepackage{caption}
\usepackage{subcaption}
\usepackage{tablefootnote}
\usepackage{amsmath}
\usepackage{amsthm}
\usepackage{mathtools}
\usepackage{algorithm}
\usepackage{algpseudocode}
\usepackage[utf8]{inputenc}
\usepackage{tikz}
\usepackage{amsfonts, amsmath, amssymb}
\usepackage{systeme, mathtools}
\usepackage{makecell}
\usepackage{listofitems} 
\usetikzlibrary{arrows.meta} 
\usepackage[outline]{contour} 
\contourlength{1.4pt}
\usetikzlibrary{positioning, arrows.meta, quotes}
\usetikzlibrary{shapes,snakes}
\usetikzlibrary{bayesnet}
\usepackage{xcolor}
\colorlet{myred}{red!80!black}
\colorlet{myblue}{blue!80!black}
\colorlet{mygreen}{green!60!black}
\colorlet{myorange}{orange!70!red!60!black}
\colorlet{mydarkred}{red!30!black}
\colorlet{mydarkblue}{blue!40!black}
\colorlet{mydarkgreen}{green!30!black}
\tikzstyle{node}=[thick,circle,draw=myblue,minimum size=22,inner sep=0.5,outer sep=0.6]
\tikzstyle{node in}=[node,green!20!black,draw=mygreen!30!black,fill=mygreen!25]
\tikzstyle{node hidden}=[node,blue!20!black,draw=myblue!30!black,fill=myblue!20]
\tikzstyle{node convol}=[node,orange!20!black,draw=myorange!30!black,fill=myorange!20]
\tikzstyle{node out}=[node,red!20!black,draw=myred!30!black,fill=myred!20]
\tikzstyle{connect}=[thick,mydarkblue] 
\tikzstyle{connect arrow}=[-{Latex[length=4,width=3.5]},thick,mydarkblue,shorten <=0.5,shorten >=1]
\tikzset{ 
  node 1/.style={node in},
  node 2/.style={node hidden},
  node 3/.style={node out},
}
\tikzset{>=latex}
\tikzstyle{plate caption} = [caption, node distance=0, inner sep=0pt,
below left=5pt and 0pt of #1.south]
\usetikzlibrary{external}
\theoremstyle{definition}
\newtheorem{theorem}{\textbf{Theorem}}[section]
\newtheorem{property}{Property}
\newtheorem{proposition}[theorem]{Proposition}

\algnewcommand{\LineComment}[1]{\State \(\triangleright\) #1}
\algnewcommand\algorithmicforeach{\textbf{for each}}
\algdef{S}[FOR]{ForEach}[1]{\algorithmicforeach\ #1\ \algorithmicdo}

\renewcommand{\r}[1]{\normalsize{\color{red}{#1}}}

\title{
DeepBern-Nets: Taming the Complexity of Certifying Neural Networks using Bernstein Polynomial Activations and Precise Bound Propagation
}

\author{%
   Haitham Khedr and Yasser Shoukry\\
  Department of Electrical Engineering and Computer Science\\
  University of California, Irvine \\
  \texttt{\{hkhedr, yshoukry\}@uci.edu} 
}

\begin{document}

\maketitle

\begin{abstract}

Formal certification of Neural Networks (NNs) is crucial for ensuring their safety, fairness, and robustness. Unfortunately, on the one hand, sound and complete certification algorithms of ReLU-based NNs do not scale to large-scale NNs. On the other hand, incomplete certification algorithms---based on propagating input domain bounds to bound the outputs of the NN---are easier to compute, but they result in loose bounds that deteriorate with the depth of NN, which diminishes their effectiveness. In this paper, we ask the following question; can we replace the ReLU activation function with one that opens the door to incomplete certification algorithms that are easy to compute but can produce tight bounds on the NN's outputs? We introduce DeepBern-Nets, a class of NNs with activation functions based on Bernstein polynomials instead of the commonly used ReLU activation. Bernstein polynomials are smooth and differentiable functions with desirable properties such as the so-called range enclosure and subdivision properties. We design a novel Interval Bound Propagation (IBP) algorithm, called Bern-IBP, to efficiently compute tight bounds on DeepBern-Nets outputs. Our approach leverages the properties of Bernstein polynomials to improve the tractability of neural network certification tasks while maintaining the accuracy of the trained networks. We conduct comprehensive experiments in adversarial robustness and reachability analysis settings to assess the effectiveness of the proposed Bernstein polynomial activation in enhancing the certification process. Our proposed framework achieves high certified accuracy for adversarially-trained NNs, which is often a challenging task for certifiers of ReLU-based NNs. Moreover, using Bern-IBP bounds for certified training results in NNs with state-of-the-art certified accuracy compared to ReLU networks. This work establishes Bernstein polynomial activation as a promising alternative for improving neural network certification tasks across various NNs applications. The code for DeepBern-Nets is publicly available\footnote{\url{https://github.com/rcpsl/DeepBern-Nets}}.

\end{abstract}


\section{Introduction}

Deep neural networks (NNs) have revolutionized numerous fields with their remarkable performance on various tasks, ranging from computer vision and natural language processing to healthcare and robotics. As these networks become integral components of critical systems, ensuring their safety, security, fairness, and robustness is essential. It is unsurprising, then, the growing interest in the field of certified machine learning, which resulted in NNs with enhanced levels of robustness to adversarial inputs ~\cite{goodfellow2014explaining,kurakin2016adversarial,song2018physical,szegedy2013intriguing}, fairness~\cite{zhang2018mitigating,xu2018fairgan,mehrabi2021survey,khedr2022certifair}, and correctness \cite{yang2019correctness}.

While certifying the robustness, fairness, and correctness of NNs with respect to formal properties is shown to be NP-hard \cite{KatzReluplexEfficientSMT2017a}, state-of-the-art certifiers rely on computing upper/lower bounds on the output of the NN and its intermediate layers~\cite{wang2021beta,khedr2021peregrinn,ferraricomplete, bak2021,henriksen2021deepsplit}.
Accurate bounds can significantly reduce the complexity and computational effort required during the certification process, facilitating more efficient and dependable evaluations of the network's behavior in diverse and challenging scenarios. Moreover, computing such bounds has opened the door for a new set of ``certified training'' algorithms ~\cite{zhang2022rethinking,lyu2021towards,muller2022certified} where these bounds are used as a regularizer that penalizes the worst-case violation of robustness or fairness, which leads to training NNs with favorable properties.
While computing such lower/upper bounds is crucial, current techniques in computing lower/upper bounds on the NN outputs are either computationally efficient but result in loose lower/upper bounds or compute tight bounds but are computationally expensive. In this paper,  we are interested in algorithms that can be both computationally efficient and lead to tight bounds.


This work follows a Design-for-Certifiability approach where we ask the question; can we replace the ReLU activation function with one that allows us to compute tight upper/lower bounds efficiently? Introducing such novel activation functions designed with certifiability in mind makes it possible to create NNs that are easier to analyze and certify during their training. Our contributions in this paper can be summarized as follows:


\begin{enumerate}
\item We introduce DeepBern-Nets, a NN architecture with a new activation function based on Bernstein polynomials. Our primary motivation is to shift some of the computational efforts from the certification phase to the training phase. By employing this approach, we can train NNs with known output (and intermediate) bounds for a predetermined input domain which can accelerate the certification process. 

\item We present Bern-IBP, an Interval Bound Propagation (IBP) algorithm that computes tight bounds of DeepBern-Nets leading to an efficient certifier.

\item We show that Bern-IBP can certify the adversarial robustness of adversarially-trained DeepBern-Nets on MNIST and CIFAR-10 datasets even with large architectures with millions of parameters. This is unlike state-of-the-art certifiers for ReLU networks, which often fail to certify robustness for adversarially-trained ReLU NNs.

\item We show that employing Bern-IBP during the training of DeepBern-Nets yields high certified robustness on the MNIST and CIFAR-10 datasets with robustness levels that are comparable---or in many cases surpassing---the performance of the most robust ReLU-based NNs reported in the SOK benchmark.
\end{enumerate}

We believe that our framework, DeepBern-Nets and Bern-IBP, enables more reliable guarantees on NN behavior and contributes to the ongoing efforts to create safer and more secure NN-based systems, which is crucial for the broader deployment of deep learning in real-world applications.



\section{DeepBern-Nets: Deep Bernstein Polynomial Networks}
\subsection{Bernstein polynomials preliminaries}
Bernstein polynomials form a basis for the space of polynomials on a closed interval \cite{Farouki2012}. These polynomials have been widely used in various fields, such as computer-aided geometric design \cite{Farouki2012}, approximation theory \cite{qian2011}, and numerical analysis \cite{farouki1987}, due to their unique properties and intuitive representation of functions.
A general polynomial of degree $n$ in Bernstein form on the interval $[l, u]$ can be represented as:
\begin{equation}
\label{eq:bern_form}
P_n^{[l,u]}(x) = \sum_{k=0}^n c_k b_{n,k}^{[l,u]}(x), \qquad x \in [l,u]
\end{equation}
where $c_k \in \mathbb{R}$ are the coefficients associated with the Bernstein basis $b_{n,k}^{[l,u]}(x)$, defined as:
\begin{equation}
\label{eq:bern_basis}
b_{n,k}^{[l,u]}(x) = \frac{\binom{n}{k}}{(u-l)^n}(x-l)^k (u - x)^{n - k},
\end{equation}
with $\binom{n}{k}$ denoting the binomial coefficient. The Bernstein coefficients $c_k$ determine the shape and properties of the polynomial $P_n^{[l,u]}(x)$ on the interval $[l, u]$. It is important to note that unlike polynomials represented in power basis form, the representation of a polynomial in Bernstein form depends on the domain of interest $[l, u]$ as shown in equation \ref{eq:bern_form}.

\begin{figure}[t]
    \centering
    \begin{subfigure}{0.45\textwidth}
    \includegraphics[width=\textwidth]{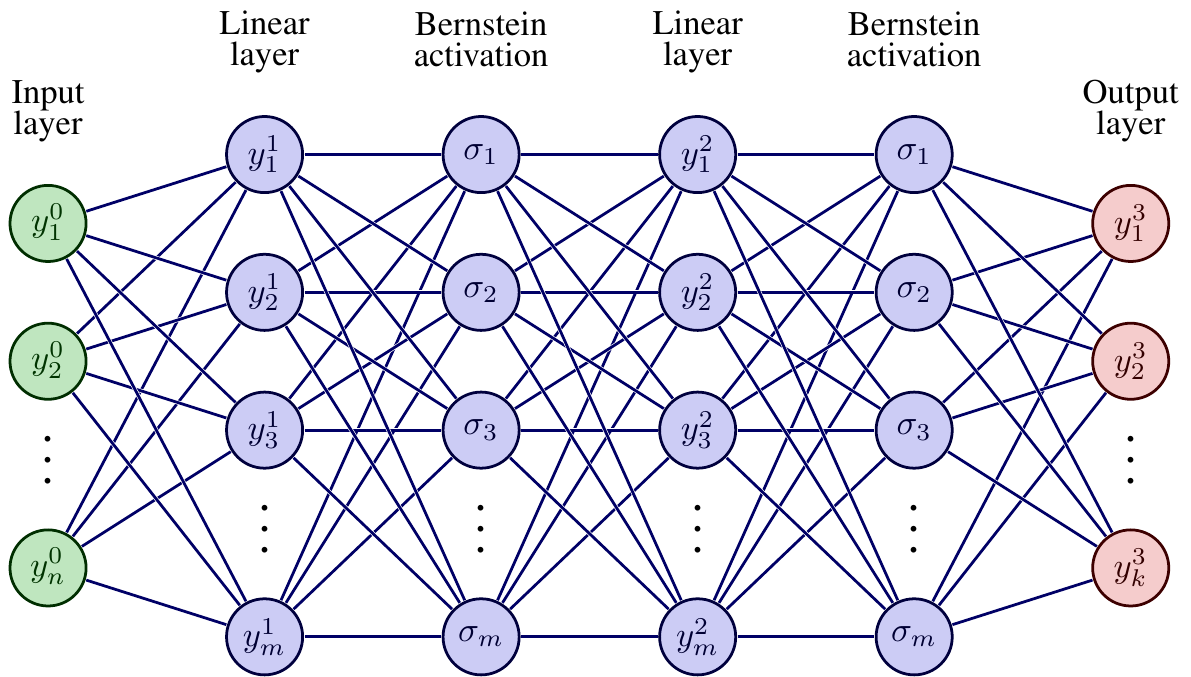}
    \end{subfigure}
    \begin{subfigure}{0.45\textwidth}
    \includegraphics[width=\textwidth, trim = 0cm 0cm 1cm 0cm]{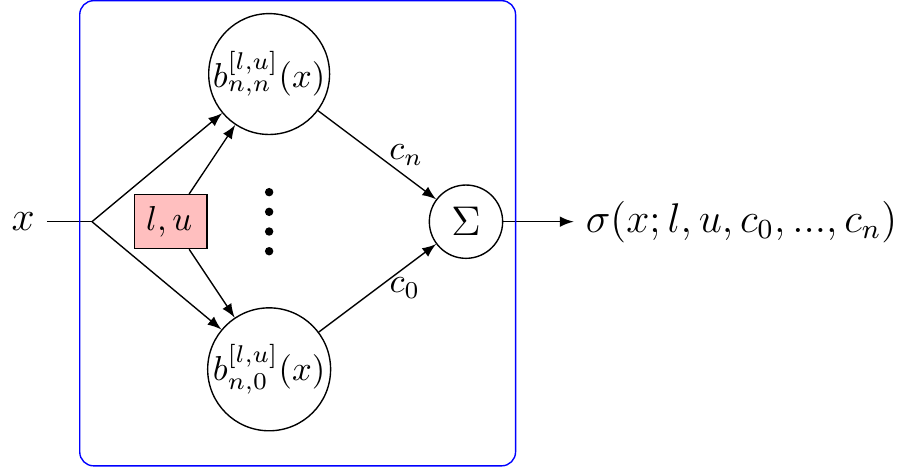}
    \end{subfigure}
    \caption{(Left) shows the structure of a DeepBern-Nets with two hidden layers. DeepBern-Nets are similar to Feed Forward NNs except that the activation function is a Bernstein polynomial. (Right) shows a simplified computational graph of a degree $n$ Bernstein activation. The Bernstein basis is evaluated at the input $x$ using $l$ and $u$ computed during training, and the output is then computed as a linear combination of the basis functions weighted by the learnable Bernstein coefficients $c_k$.} 
    \label{fig:bern_NN} 
\end{figure}

\subsection{Neural Networks with Bernstein activation functions}
We propose using Bernstein polynomials as non-linear activation functions $\sigma$ in feed-forward NNs. We call such NNs as DeepBern-Nets. Like feed-forward NNs, DeepBern-Nets consist of multiple layers, each consisting of linear weights followed by non-linear activation functions. Unlike conventional activation functions (e.g., ReLU, sigmoid, tanh, ..), Bernstein-based activation functions are parametrized with learnable Bernstein coefficients $\boldsymbol{c} = c_0, \ldots, c_n$, i.e.,
\begin{equation}
\sigma(x;l,u,\boldsymbol{c}) = \sum_{k=0}^n c_k b_{n,k}^{[l,u]}(x), \qquad x \in [l,u] \text{,}
\end{equation}
where $x$ is the input to the neuron activation, and the polynomial degree $n$ is an additional hyper-parameter of the Bernstein activation and can be chosen differently for each neuron. Figure \ref{fig:bern_NN} shows a simplified computational graph of the Bernstein activation and how it is used to replace conventional activation functions.

\paragraph{Training of DeepBern-Nets.}Since Bernstein polynomials are defined on a specific domain (equation \ref{eq:bern_basis}), we need to determine the lower and upper bounds ($\boldsymbol{l}^{(k)}$ and $\boldsymbol{u}^{(k)}$) of the inputs to the Bernstein activation neurons in layer $k$, during the training of the network. To that end, we assume that the input domain $\mathcal{D}$ is bounded with the lower and upper bounds (denoted as $\boldsymbol{l}^{(0)}$ and $\boldsymbol{u}^{(0)}$, respectively) known during training. 
We emphasize that our assumption that $\mathcal{D}$ is bounded and known is not conservative, as the input to the NN can always be normalized to $[0, 1]$, for example.

\begin{algorithm}[!b]
    \caption{Training step of an L-layer DeepBern-Net $\mathcal{NN}$}
    \label{alg:training_step}
\begin{algorithmic}[1]
\State Given: Training Batch  ($\mathcal{X}, \boldsymbol{t}$) and input bounds $[\boldsymbol{l}^{(0)}$, $\boldsymbol{u}^{(0)}]$ 
\State Initialize all parameters
\State Set the learning rate $\alpha$
\LineComment{\textcolor{red}{Forward propagation}}
\State Set $\boldsymbol{y}^{(0)} = \mathcal{X}$
\State Set $\mathcal{B}^{(0)} = [\boldsymbol{l}^{(0)}$, $\boldsymbol{u}^{(0)}]$

\For{$i$ = $1$....$L$}
    \If{$layer$ $i$ is Bernstein activation}
        \State $\boldsymbol{l}^{(i)}, \boldsymbol{u}^{(i)} \gets \mathcal{B}^{(i-1)}$ \Comment{\textcolor{red}{Store Input bounds of the Bernstein layer}}
        \ForEach{neuron $z$ in $layer$ $i$}
            \State Let $\boldsymbol{c^{(i)}_{z}}$ be the Bernstein coefficients for neuron $z$ of the $i$-th layer
            \State $\mathcal{B}^{(i)}_z \gets [\min\limits_j{c^{(i)}_{zj}}, \max\limits_j{c^{(i)}_{zj}}]$ \label{alg:train:bernibp}
        \EndFor
        \State $\mathcal{B}^{(i)} \gets [\mathcal{B}_0^{(i)}, \mathcal{B}_1^{(i)},...,\mathcal{B}_m^{(i)}]$ \Comment{$m$ denotes the number of neurons in layer $i$}
    \Else
        \State $\mathcal{B}^{(i)} \gets $IBP$(\mathcal{B}^{(i-1)})$  \label{line:ibp}
    \EndIf
    \State $\boldsymbol{y}^{(i)} \gets  \text{forward}(\boldsymbol{y}^{(i-1)})$\Comment{\textcolor{red}{Regular forward step}}
\EndFor
\LineComment{\textcolor{red}{Backpropagation}}
\State Compute the loss function: $\mathcal{L}(\boldsymbol{y}^{(L)}, \boldsymbol{t})$
\State Compute the gradients with respect to all model parameters (including Bernstein coefficients)
\ForEach {Parameter $\theta$} \Comment{\r{Weights, biases, and Bernstein coefficients $c_k$}}
\State $\theta \gets  \theta - \alpha \nabla_{\theta} \mathcal{L}$
\EndFor

\end{algorithmic}
\end{algorithm}

Using the bounds on the input domain $\boldsymbol{l}^{(0)}$ and $\boldsymbol{u}^{(0)}$ and the learnable parameters of the NNs (i.e., weights of the linear layers and the Bernstein coefficients $\boldsymbol{c}$ for each neuron), we will 
update the bounds $\boldsymbol{l}^{(k)}$ and $\boldsymbol{u}^{(k)}$ with each step of training by propagating $\boldsymbol{l}^{(0)}$ and $\boldsymbol{u}^{(0)}$ through all the layers in the network. 
Unlike conventional non-linear activation functions where symbolic bound propagation relies on linear relaxation techniques ~\cite{wang2018efficient, wang2018formal}, the Bernstein polynomial enclosure property allows us to bound the output of an $n$-th order Bernstein activation in $\mathcal{O}(n)$ operations (Algorithm \ref{alg:training_step}-line \ref{alg:train:bernibp}). We start by reviewing the enclosure property of Bernstein polynomials as follows.

\begin{property}[Enclosure of Range~\cite{titi2019}]
\label{prop:enclosure}
The enclosure property of Bernstein polynomials 
states that for a given polynomial $P_n^{[l,u]}(x)$ of degree $n$ in Bernstein form on an interval $[l, u]$, the polynomial lies within the convex hull of its Bernstein coefficients. In other words, the Bernstein polynomial is bounded by the minimum and maximum values of its coefficients $c_k$ regardless of the input $x$.

\begin{equation}
\min_{0 \leq k \leq n} c_k \leq P_n^{[l,u]}(x) \leq \max_{0 \leq k \leq n} c_k, \qquad \forall x \in [l, u].
\end{equation}
\end{property}


Algorithm~\ref{alg:training_step} outlines how to use the enclosure property to propagate the bounds from one layer to another for a single training step in an L-layer DeepBern-Net. In contrast to normal training, we calculate the worst-case bounds for the inputs to all Bernstein layers by propagating the bounds from the previous layers. Such bound propagation can be done for linear layers using interval arithmetic ~\cite{liu2021algorithms}---referred to in Algorithm \ref{alg:training_step}-line~\ref{line:ibp} as Interval Bound Propagation (IBP)---or using Property~\ref{prop:enclosure} for Bernstein layers (Algorithm \ref{alg:training_step}-Line~\ref{alg:train:bernibp}). We store the resulting bounds for each Bernstein activation function. Then, we perform the regular forward step. The parameters are then updated using vanilla backpropagation, just like conventional NNs. During inference, we directly use the stored layer-wise bounds $\boldsymbol{l}^{(k)}$ and $\boldsymbol{u}^{(k)}$ (computed during training) to propagate any input through the network. 
%
In Appendix \ref{app:timing}, we show that the overhead of computing the bounds $\boldsymbol{l}^{(k)}$ and $\boldsymbol{u}^{(k)}$ during training adds between $0.2\times$ to $5\times$ overhead for the training, depending on the order $n$ of the Bernstein activation function and the size of the network.

\paragraph{Stable training of DeepBern-Nets.} Using polynomials as activation functions in deep NNs has attracted several researchers' attention in recent years~\cite{wang2022,gottemukkula2020polynomial}. A major drawback of using polynomials of arbitrary order is their unstable behavior during training due to exploding gradients--which is prominent with the increase in order~\cite{gottemukkula2020polynomial}. In particular, for a general $n$th order polynomial in power series $f_n(x) = w_0 + w_1 x + \ldots + w_n x^n$, its derivative is $df_n(x)/dx = w_1 + \ldots + n w_n x^{n-1}$. Hence training a deep NN with multiple polynomial activation functions suffers from exploding gradients as the gradient scales exponentially with the increase in the order $n$ for $x > 1$.

Luckily, and thanks to the unique properties of Bernstein polynomials, DeepBern-Net does not suffer from such a limitation as captured in the next result, whose proof is given in Appendix~\ref{app:proof}. 

\begin{proposition}
    \label{prop:gradient}
    Consider the Bernstein activation function $\sigma(x;l,u,\boldsymbol{c})$ of arbitrary order $n$. The following holds:
    \begin{enumerate}
        \item $\left\vert \frac{d}{dx} \sigma(x;l,u,\boldsymbol{c}) \right\vert \le 2 n \max_{k \in \{0,\ldots,n\}} \vert c_k \vert$,
        \item $\left\vert \frac{d}{dc_i} \sigma(x;l,u,\boldsymbol{c}) \right\vert \le 1 \quad$  for all $ i \in \{0, \ldots,  n\}$.
   \end{enumerate}
    
\end{proposition}

Proposition~\ref{prop:gradient} ensures that the gradients of the proposed Bernstein-based activation function depend only on the value of the learnable parameters $\boldsymbol{c} = (c_0, \ldots, c_n)$. Hence, the gradients do not explode for $x > 1$. This feature is not enjoyed by the polynomial activation functions in~\cite{gottemukkula2020polynomial} and leads to better stable training properties when the Bernstein polynomials are used as activation functions. Moreover, one can control these gradients by adding a regularizer--to the objective function--that penalizes high values of $c_k$, which is common for other learnable parameters, i.e., weights of the linear layer. Proof of Proposition \ref{prop:gradient} is in Appendix \ref{app:proof}



\section{Bern-IBP: Certification using Bernstein Interval Bound Propagation}

\subsection{Certification of global properties using Bern-IBP}
\label{sec:cert_glob}


We consider the certification of global properties of NNs. Global properties need to be held true for the entire input domain $\mathcal{D}$ of the network. For simplicity of presentation, we will assume that the global property we want to prove takes the following form: 
\begin{align}
\label{eq:global_prop}
\forall \boldsymbol{y^{(0)}} \in \mathcal{D} \Longrightarrow y^{(L)} = \mathcal{NN}(\boldsymbol{y}^{(0)}) > 0 
\end{align}
where $y^{(L)}$ is a scalar output and $\mathcal{NN}$ is the NN of interest. Examples of such global properties include the stability of NN-controlled systems~\cite{wu2022} as well as global individual fairness~\cite{khedr2022certifair}.


In this paper, we focus on the incomplete certification of such properties. In particular, we certify properties of the form~\eqref{eq:global_prop} by checking the lower/upper bounds of the NN. To that end, we define the lower  $\mathcal{L}$ and upper $\mathcal{U}$ bounds of the NN within the domain $\mathcal{D}$ as any real numbers that satisfy:
\begin{align}
    &\mathcal{L}\left(\mathcal{NN}(\boldsymbol{y}^{(0)}), \mathcal{D} \right) \le \min_{\boldsymbol{y}^{(0)} \in \mathcal{D}} \mathcal{NN}(\boldsymbol{y}^{(0)}) ,\qquad
    \mathcal{U}\left(\mathcal{NN}(\boldsymbol{y}^{(0)}), \mathcal{D} \right) \ge \max_{\boldsymbol{y}^{(0)} \in \mathcal{D}} \mathcal{NN}(\boldsymbol{y}^{(0)})
\end{align}
Incomplete certification of~\eqref{eq:global_prop} is equivalent to checking if $\mathcal{L}\left(\mathcal{NN}(\boldsymbol{y}^{(0)}), \mathcal{D} \right) > 0$. Thanks to the Enclosure of Range (Property~\ref{prop:enclosure}) of DeepBern-Nets, one can check the condition $\mathcal{L}\left(\mathcal{NN}(\boldsymbol{y}^{(0)}), \mathcal{D} \right) > 0$ in constant time, i.e., $\mathcal{O}(1)$, by simply checking the minimum Bernstein coefficients of the output layer.





\subsection{Certification of local properties using Bern-IBP}
\label{sec:cert_local}

Local properties of NNs are the ones that need to be held for subsets $S$ of the input domain $\mathcal{D}$, i.e.,
\begin{align}
\label{eq:local_prop}
\forall \boldsymbol{y^{(0)}} \in S \subset \mathcal{D} \Longrightarrow y^{(L)} = \mathcal{NN}(\boldsymbol{y}^{(0)}) > 0 
\end{align}
Examples of local properties include adversarial robustness and the safety of NN-controlled vehicles~\cite{sun2019formal, kochdumper2023provably, santa2022nnlander}. Similar to global properties, we are interested in incomplete certification by checking whether $\mathcal{L}\left(\mathcal{NN}(\boldsymbol{y}^{(0)}), S \right) > 0$.

The output bounds stored in the Bernstein activation functions are the worst-case bounds for the entire input domain $\mathcal{D}$. However, for certifying local properties over $S \subset \mathcal{D}$, we need to refine these output bounds on the given sub-region $S$. To that end, for a Bernstein activation layer $k$ with input bounds [$\boldsymbol{l}^{(k)}$, $\boldsymbol{u}^{(k)}$] (computed and stored during training), we can obtain tighter output bounds thanks to the following subdivision property of Bernstein polynomials. 
\begin{property}[Subdivision~\cite{titi2019}]
\label{prop:subdivision}
Given a Bernstein polynomial $P_n^{[l,u]}(x)$ of degree $n$ on the interval $[l, u]$, the coefficients of the same polynomial on subintervals $[l, \alpha]$ and $[\alpha, u]$ with $\alpha \in [l, u]$ can be computed as follows. First, compute the intermediate coefficients $c_j^k$ for $k = 0,...,n$ and $j = k,...,n$
\begin{align}
\centering
c_j^k &= 
\left\{
	\begin{array}{ll}
		c_j  & \mbox{if } k = 0 \\
		(1-\tau)c_{j-1}^{k-1} + \tau c_j^{k-1} & \mbox{if } k > 0
	\end{array}
\right., \quad 
c_i' = c_i^i    \quad 
c_i'' = c_n^{n-i} \quad i=0\ldots n \nonumber, 
\end{align}
where $\tau=\frac{\alpha - l}{u-l}$. Next, the
polynomials defined on each of the subintervals $[l, \alpha]$ and $[\alpha, u]$ are:
\begin{align*}
    P_n^{[l,\alpha]}(x) &= \sum_{k=0}^n c_k' b_{n,k}^{[l,\alpha]}(x), \qquad
    P_n^{[\alpha,u]}(x) = \sum_{k=0}^n c_k'' b_{n,k}^{[\alpha,u]}(x).
\end{align*}
\end{property} 

Indeed, we can apply the Subdivision property twice to compute the coefficients of the polynomial $P_n^{[\alpha,\beta]}$. Computing the coefficients on the subintervals allows us to tightly bound the polynomial using property \ref{prop:enclosure}. Therefore, given a DeepBern-Net trained on $\mathcal{D} = [\boldsymbol{l}^{(0)}$, $\boldsymbol{u}^{(0)}]$, we can compute tighter bounds on the subregion $S = [\boldsymbol{\hat{l}}^{(0)}$, $\boldsymbol{\hat{u}}^{(0)}]$ by applying the subdivision property (Property \ref{prop:subdivision}) to compute the Bernstein coefficients on the sub-region $S$, and then use the enclosure property (Property \ref{prop:enclosure}) to compute tight bounds on the output of the activation equivalent to the minimum and maximum of the computed Bernstein coefficients. We do this on a layer-by-layer basis until we reach the output of the NN. Implementation details of this approach is given in Appendix~\ref{sec:implementation}.

\section{Experiments}
\textbf{Implementation:} Our framework has been developed in Python, and is designed to facilitate the training of DeepBern-Nets and certify local properties such as Adversarial Robustness and certified training. We use PyTorch \cite{paszke2019pytorch} for all neural network training tasks. To conduct our experiments, we utilized a single GeForce RTX 2080 Ti GPU in conjunction with a 24-core Intel(R) Xeon(R) CPU E5-2650 v4 @ 2.20GHz. Only 8 cores were utilized for our experiments. 
 
\subsection{Experiment 1: Certification of Adversarial Robustness}

The first experiment assesses the ability to compute tight bounds on the NN output and its implications for certifying NN properties. To that end, we use the application of adversarial robustness, where we aim to certify that a NN model is not susceptible to adversarial examples within a defined perturbation set. The results in~\cite{li2020sok, muller2022third} show that state-of-the-art IBP algorithms fail to certify the robustness of NNs trained with Projected Gradient Descent (PGD), albeit being robust, due to the excessive errors in the computed bounds, which forces designers to use computationally expensive sound and complete algorithms. Thanks to the properties of DeepBern-Nets, the bounds computed by Bern-IBP are tight enough to certify the robustness of NNs without using computationally expensive sound and complete tools. To that end, we trained several NNs using the MNIST \cite{mnist} and CIFAR-10 \cite{cifar10} datasets using PGD. We trained both Fully Connected Neural Networks (FCNN) and Convolutional Neural Networks (CNNs) on these datasets with Bernstein polynomials of orders $2,3,4,5$, and $6$. For detailed information regarding the model architectures, please refer to Appendix \ref{app:arch}. Further information about the training procedure can be found in Appendix \ref{app:exp_setup}.




\subsubsection{Formalizing adversarial robustness as a local property}
Given a NN model $\mathcal{NN}:[0,1]^d\to\mathbb{R}^o$, a concrete input $\boldsymbol{x_n}$, a target class $t$, and a perturbation parameter $\epsilon$, the adversarial robustness problem asks that the NN output be the target class $t$ for all the inputs in the set $\{\boldsymbol{x} \mid \|\boldsymbol{x} -\boldsymbol{x_n}\|_\infty \leq \epsilon\}$. In other words, a NN is robust whenever:
$$ \forall \boldsymbol{x} \in S(\boldsymbol{x_n}, \epsilon) = \{\boldsymbol{x} \mid \|\boldsymbol{x} -\boldsymbol{x_n}\|_\infty \leq \epsilon\} \quad \Longrightarrow \quad \mathcal{NN}(\boldsymbol{x})_{t} > \mathcal{NN}(\boldsymbol{x})_{i},  \; i \ne t $$
where $\mathcal{NN}(\boldsymbol{x})_{t}$ is the NN output for the target class and $\mathcal{NN}(\boldsymbol{x})_{i}$ is the NN output for any class $i$ other that $t$.
To certify the robustness of a NN, one can compute a lower bound on the adversarial robustness $\mathbb{L}_{\text{robust}}$ for all classes $i \ne t$ as:
\begin{align}
\label{eq:robustness_lb}
\centering
\mathbb{L}_{\text{robust}}(\boldsymbol{x_n}, \epsilon) &= \min_{i \ne t} \Bigg(\mathcal{L}\big(\mathcal{NN}(\boldsymbol{x})_{t}, S(\boldsymbol{x_n}, \epsilon)) - \mathcal{U}\big(\mathcal{NN}(\boldsymbol{x})_{i}, S(\boldsymbol{x_n}, \epsilon)\big) \Bigg) \\
&\le 
\min_{i \ne t} \left(
\min_{\boldsymbol{x} \in S(\boldsymbol{x_n}, \epsilon)} \mathcal{NN}(\boldsymbol{x})_{t} - \mathcal{NN}(\boldsymbol{x})_{i} \right)
\end{align}
Indeed, the NN is robust whenever $\mathbb{L}_{\text{robust}} > 0$. Nevertheless, the tightness of the bounds $\mathcal{L}(\mathcal{NN}(\boldsymbol{x})_{t}, S(\boldsymbol{x_n}, \epsilon)) $ and $\mathcal{U}(\mathcal{NN}(\boldsymbol{x})_{i}, S(\boldsymbol{x_n}, \epsilon))$ plays a significant role in the ability to certify the NN robustness. The tighter these bounds, the higher the ability to certify the NN robustness.

\subsubsection{Experiment 1.1: Tightness of output bounds - Bern-IBP vs IBP}
\label{exp:bounds}

\begin{figure}[ht]
    \centering
    \includegraphics[width=\textwidth]{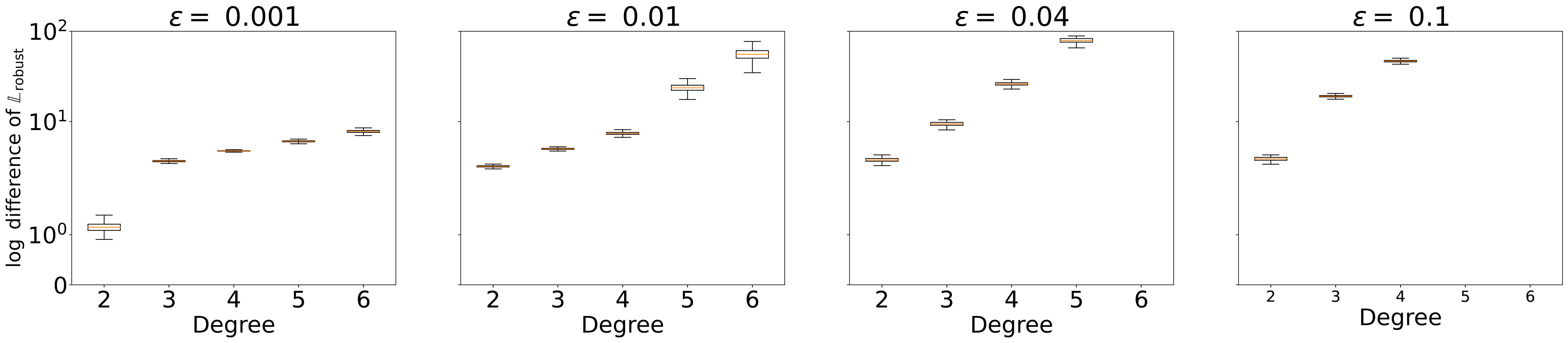}
    \caption{A visual representation of the tightness of bounds computed using Bern-IBP compared to IBP. The figure shows the $\log$ difference between $\mathbb{L}_{\text{robust}}$ computed using Bern-IBP and IBP for NNs with varying orders of and different values of $\epsilon$. The figure demonstrates the enhanced precision and scalability of the Bern-IBP method in computing tighter bounds, even for higher-order Bernstein activations and larger values of $\epsilon$, as compared to the naive IBP method.
}
    \label{fig:bern-ibp} 
\end{figure}

For each trained neural network, we compute the lower bound on robustness $\mathbb{L}_{\text{robust}}(\boldsymbol{x_n}, \epsilon)$ using Bern-IBP and using state-of-the-art Interval Bound Propagation (IBP) that does not take into account the properties of DeepBern-Nets. In particular, for this experiment, we used auto\_LiRPA\cite{wang2021beta}, a tool that is part of $\alpha\beta$-CROWN\cite{wang2021beta}---the winner of the 2022 Verification of Neural Network (VNN) competition~\cite{muller2022third}. Figure~\ref{fig:bern-ibp} shows the difference between the bound $\mathbb{L}_{\text{robust}}(\boldsymbol{x_n}, \epsilon)$ computed by Bern-IBP and the one computed by IBP using a semi-log scale. The raw data for the adversarial robustness bound $\mathbb{L}_{\text{robust}}(\boldsymbol{x_n}, \epsilon)$ for both Bern-IBP and IBP is given in Appendix~\ref{app:lower_bounds}.

The results presented in Figure~\ref{fig:bern-ibp} clearly demonstrate that  Bern-IBP yields significantly tighter bounds in comparison to IBP. Figure~\ref{fig:bern-ibp} also shows that for all values of $\epsilon$, the bounds computed using IBP become exponentially looser as the order of the Bernstein activations increase, unlike the bounds computed with Bern-IBP, which remain precise even for higher-order Bernstein activations or larger values of $\epsilon$. The raw data in Appendix~\ref{app:lower_bounds} provide a clearer view on the superiority of computing $\mathbb{L}_{\text{robust}}(\boldsymbol{x_n}, \epsilon)$ using Bern-IBP compared to IBP.




\subsubsection{Experiment 1.2: Certification of Adversarial Robustness using Bern-IBP}
\label{sec:robustness}

\begin{table}[!b]
\centering
\caption{A comparison of certified accuracy and verification time for neural networks with Bernstein polynomial activations using both IBP and Bern-IBP methods and varying values of $\epsilon$. The table also presents the upper bound on certified accuracy calculated using a 100-step PGD attack. The results highlight the superior performance of Bern-IBP in certifying robustness properties compared to IBP.
}
\label{tab:cert_acc}
\resizebox{\textwidth}{!}{%
\begin{tabular}{cccc|cc|cc|c}
\toprule
\multirow{2}{*}{Dataset} &
  \multirow{2}{*}{\makecell{Model \\ (\# of params)}} &
  \multirow{2}{*}{Test acc. (\%)} &
  \multirow{2}{*}{$\epsilon$} &
  \multicolumn{2}{c|}{IBP} &
  \multicolumn{2}{c|}{Bern-IBP} &
  U.B (PGD) \\
 &
   &
   &
   &
  Time (s) &
  \begin{tabular}[c]{@{}c@{}}Certified \\ acc. (\%)\end{tabular} &
  Time (s) &
  \begin{tabular}[c]{@{}c@{}}Certified \\ acc. (\%)\end{tabular} &
  \begin{tabular}[c]{@{}c@{}}Certified \\ acc. (\%)\end{tabular}
  \\ \hline
\multirow{6}{*}{MNIST}   & \multirow{3}{*}{\makecell{CNNa\_4 \\ (190,426)}} & \multirow{3}{*}{97.229} & 0.01  & 3.45 & 0 & 1.43 & 88.69 & 95.97 \\
                         &                          &                         & 0.03  & 3.41 & 0 & 1.42 & 72.12 & 92.53 \\
                         &                          &                         & 0.1   & 3.26 & 0 & 1.39 & 65.22 & 75.27 \\
                         & \multirow{3}{*}{\makecell{CNNb\_2 \\ (905,882)}} & \multirow{3}{*}{97.14}  & 0.01  & 4.38 & 0 & 2.07 & 80.21 & 95.42 \\
                         &                          &                         & 0.03  & 4.58 & 0 & 2.11 & 56.49 & 90.57 \\
                         &                          &                         & 0.1   & 4.61 & 0 & 1.97 & 72.35 & 78.6  \\ \hline
\multirow{4}{*}{CIFAR-10} & \multirow{2}{*}{\makecell{CNNa\_6 \\ (258,626)}} & \multirow{2}{*}{46.77}  & 1/255 & 3.29 & 0 & 1.82 & 27.74 & 33.53 \\
                         &                          &                         & 2/255 & 3.25 & 0 & 1.83 & 33.49 & 35.81 \\
                         & \multirow{2}{*}{\makecell{CNNb\_4\\ (1,235,994)}} & \multirow{2}{*}{54.66}  & 1/255 & 5.17 & 0 & 4.45 & 28.55 & 42.86 \\
                         &                          &                         & 2/255 & 5.14 & 0 & 4.33 & 14.7  & 36.73 \\
\bottomrule                         
\end{tabular}%
}
\end{table}

Next, we show that the superior precision of bounds calculated using Bern-IBP can lead to efficient certification of adversarial robustness. Here, we define the certified accuracy of the NN as the percentage of the data points (in the test dataset) for which an adversarial input can not change the class (the output of the NN). Table~\ref{tab:cert_acc} contrasts the certified accuracy for the adversarially-trained (using 100-step PGD) DeepBern-Nets of orders 2, 4, and 6, using both IBP and Bern-IBP methods and varying values of $\epsilon$. As observed by the table, IBP fails to certify the robustness of all the NNs. On the other hand, Bern-IBP achieved high certified accuracy for all the NNs with varying values of $\epsilon$. 
Finally, we use the methodology reported in~\cite{wang2021beta} to upper bound the certified accuracy using  100-step PGD attack. 





It is essential to mention that IBP's inability to certify the robustness of NNs is not unique to DeepBern-Nets. In particular, as shown in~\cite{li2020sok, muller2022third}, most certifiers struggle to certify the robustness of ReLU NNs when trained with PGD. This suggests the power of DeepBern-Nets, which can be efficiently certified---in a few seconds even for NNs with millions of parameters, as shown in Table~\ref{tab:cert_acc}---using incomplete certifiers thanks to the ability of Bern-IBP to compute tight bounds.


\subsection{Experiment 2: Certified training using Bern-IBP}

In this experiment, we demonstrate that the tight bounds calculated by Bern-IBP can be utilized for certified training, achieving state-of-the-art results. Although a direct comparison with methods from certified training literature is not feasible due to the use of Bernstein polynomial activations instead of ReLU activations, we provide a comparison with state-of-the-art certified accuracy results from the SOK benchmark~\cite{li2020sok} to study how effectively can Bern-IBP be utilized for certified training. We trained neural networks with the same architectures as those in the benchmark to maintain a similar number of parameters, with the polynomial order serving as an additional hyperparameter. The training objective adheres to the certified training literature \cite{zhang2019towards}, incorporating the bound on the robustness loss in the objective as follows:
\begin{equation}
    \min_{\theta} \mathop{\mathbb{E}}_{(\boldsymbol{x},y) \in (X,Y)} \bigg[(1-\lambda)\mathcal{L}_{\text{CE}}(\mathcal{NN}_\theta(\boldsymbol{x}),y;\theta) + 
    \lambda \mathcal{L}_{\text{RCE}}(S(\boldsymbol{x}, \epsilon),y;\theta))\bigg],
    \label{eq:train_loss}
\end{equation}
where $\boldsymbol{x}$ is a data point, $y$ is the ground truth label, $\lambda \in [0,1]$ is a weight to control the certified training regularization, $\mathcal{L}_{\text{CE}}$ is the cross-entropy loss, $\theta$ is the NN parameters, and $\mathcal{L}_{\text{RCE}}$ is computed by evaluating $\mathcal{L}_{\text{CE}}$ on the upper bound of the logit differences computed\cite{zhang2019towards} using a bounding method.

For DeepBern-Nets, $\mathcal{L}_{\text{RCE}}$ is computed using Bern-IBP during training, while the networks in the SOK benchmark are trained using CROWN-IBP \cite{zhang2019towards}. Table~\ref{tab:robustness} illustrates that employing Bern-IBP bounds for certified training yields state-of-the-art certified accuracy (certified with Bern-IBP) on these datasets, comparable to---or in many cases surpassing---the performance of ReLU networks. The primary advantage of using Bern-IBP lies in its ability to compute highly precise bounds using a computationally cheap method, unlike the more sophisticated bounding methods for ReLU networks, such as $\alpha$-Crown. For more details about the exact architecture of the NNs, please refer to Appendix~\ref{app:arch}
\begin{table}[!b]
\centering
\caption{A comparison of certified accuracy for NNs with Bernstein polynomial activations versus ReLU NNs as in the SOK benchmark \cite{li2020sok}. The certified accuracy is computed using Bern-IBP for NNs with polynomial activations, and the method yielding highest certified accuracy as reported in SOK for ReLU NNs.
The table highlights the effectiveness of Bern-IBP in achieving competitive certification while utilizing a very computationally cheap method for tight bound computation.
}
\label{tab:robustness}
\resizebox{\textwidth}{!}{%
\begin{tabular}{c|cccc|cccc}
\toprule

\multirow{2}{*}{Model} &
  \multicolumn{4}{c|}{MNIST Certified acc. (\%)} &
  \multicolumn{4}{c}{CIFAR-10 Certified acc. (\%)} \\
 &
  \multicolumn{2}{c|}{$\epsilon = 0.1$} &
  \multicolumn{2}{c|}{$\epsilon = 0.3$} &
  \multicolumn{2}{c|}{$\epsilon = 2/255$} &
  \multicolumn{2}{c}{$\epsilon = 8/255$} \\
 &
  \makecell{DeepBern-Net \\(\%)} &
  \multicolumn{1}{c|}{\makecell{SOK \\ (\%)}} &
  \makecell{DeepBern-Net \\(\%)} &
  \makecell{SOK \\ (\%)} &
  \makecell{DeepBern-Net \\(\%)} &
  \multicolumn{1}{c|}{\makecell{SOK \\ (\%)}} &
  \makecell{DeepBern-Net \\(\%)} &
  \makecell{SOK \\ (\%)} \\ \hline
FCNNa & \textbf{72} & \multicolumn{1}{c|}{68} & \textbf{31} & 25 & \textbf{38} & \multicolumn{1}{c|}{33} & \textbf{28} & 27 \\
FCNNb & \textbf{86} & \multicolumn{1}{c|}{85} & \textbf{57} & 54 & \textbf{39} & \multicolumn{1}{c|}{37} & \textbf{26} & 25 \\
FCNNc & \textbf{80} & \multicolumn{1}{c|}{80} & \textbf{51} & 22 & \textbf{36} & \multicolumn{1}{c|}{32} & \textbf{31} & 30 \\
CNNa  & \textbf{95} & \multicolumn{1}{c|}{95} & 82 & \textbf{88} & 45 & \multicolumn{1}{c|}{\textbf{46}} & 31 & \textbf{34} \\
CNNb  & \textbf{95} & \multicolumn{1}{c|}{94} & 77 & \textbf{85} & \textbf{49} & \multicolumn{1}{c|}{49} & \textbf{37} & 35 \\
CNNc  & 87 & \multicolumn{1}{c|}{\textbf{89}} & 72 & \textbf{87} & 38 & \multicolumn{1}{c|}{\textbf{51}} & 32 & \textbf{38}\\
\bottomrule
\end{tabular}%
}
\end{table}

\subsection{Experiment 3: Tight reachability analysis of NN-controlled Quadrotor using Bern-IBP}

In this experiment, we study the application-level impact of using Bernstein polynomial activations in comparison to ReLU activations with respect to the tightness of reachable sets in the context of safety-critical applications. Specifically, we consider a 6D linear dynamics system  $\dot{x} = Ax + Bu$ representing a Quadrotor (used in ~\cite{everett2021reachability,hu2020reach, arch19}), controlled by a nonlinear NN controller where $u = \mathcal{NN}(x)$. To ensure a fair comparison, both sets of networks are trained on the same datasets, using the same architectures and training procedures. The only difference between the two sets of networks is the activation function used (ReLU vs. Bernstein polynomial).

\begin{figure}
    \centering
    \begin{subfigure}{0.49\textwidth}
    \includegraphics[height=4cm]{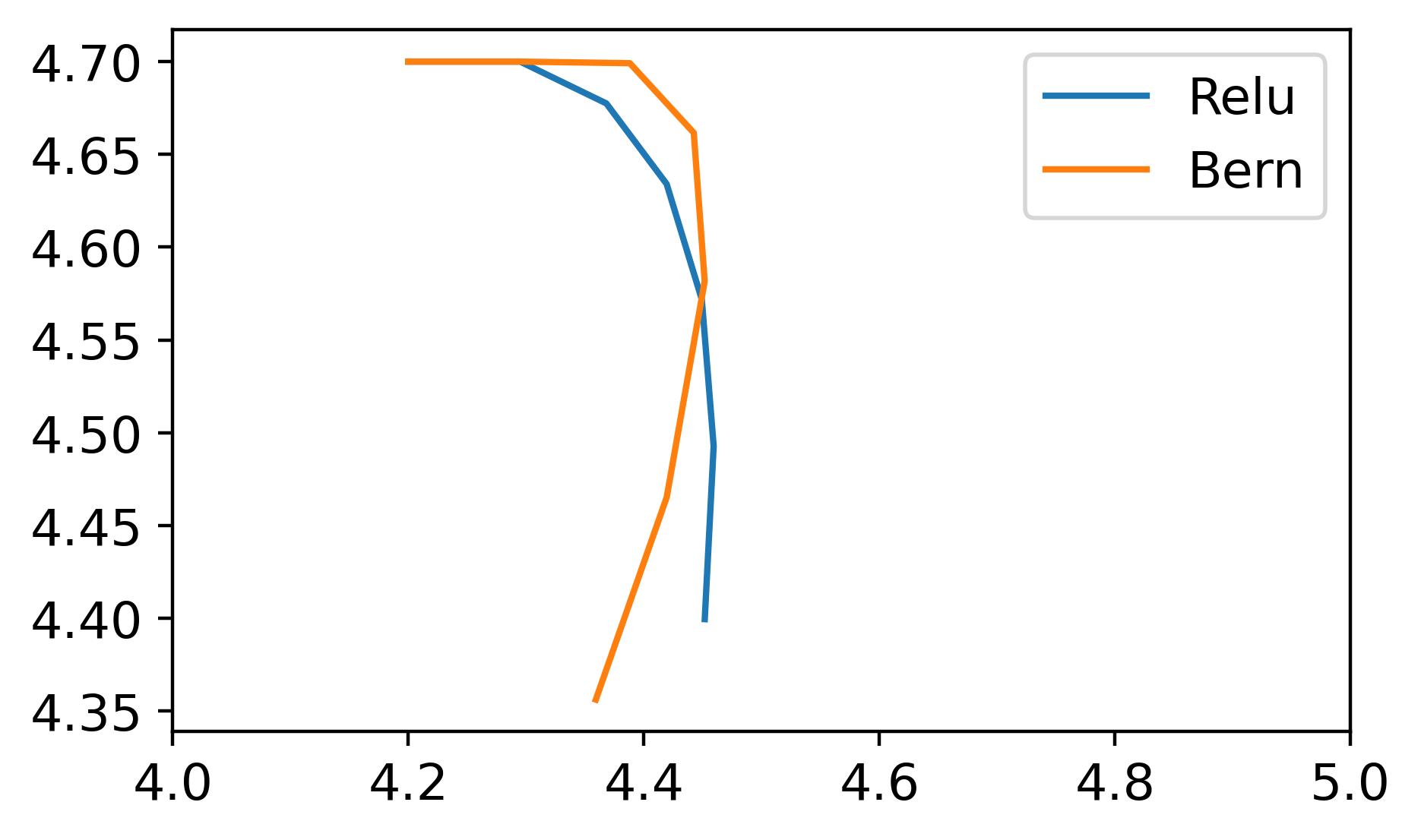}
    \end{subfigure}
    \begin{subfigure}{0.45\textwidth}
    \includegraphics[height=4cm]{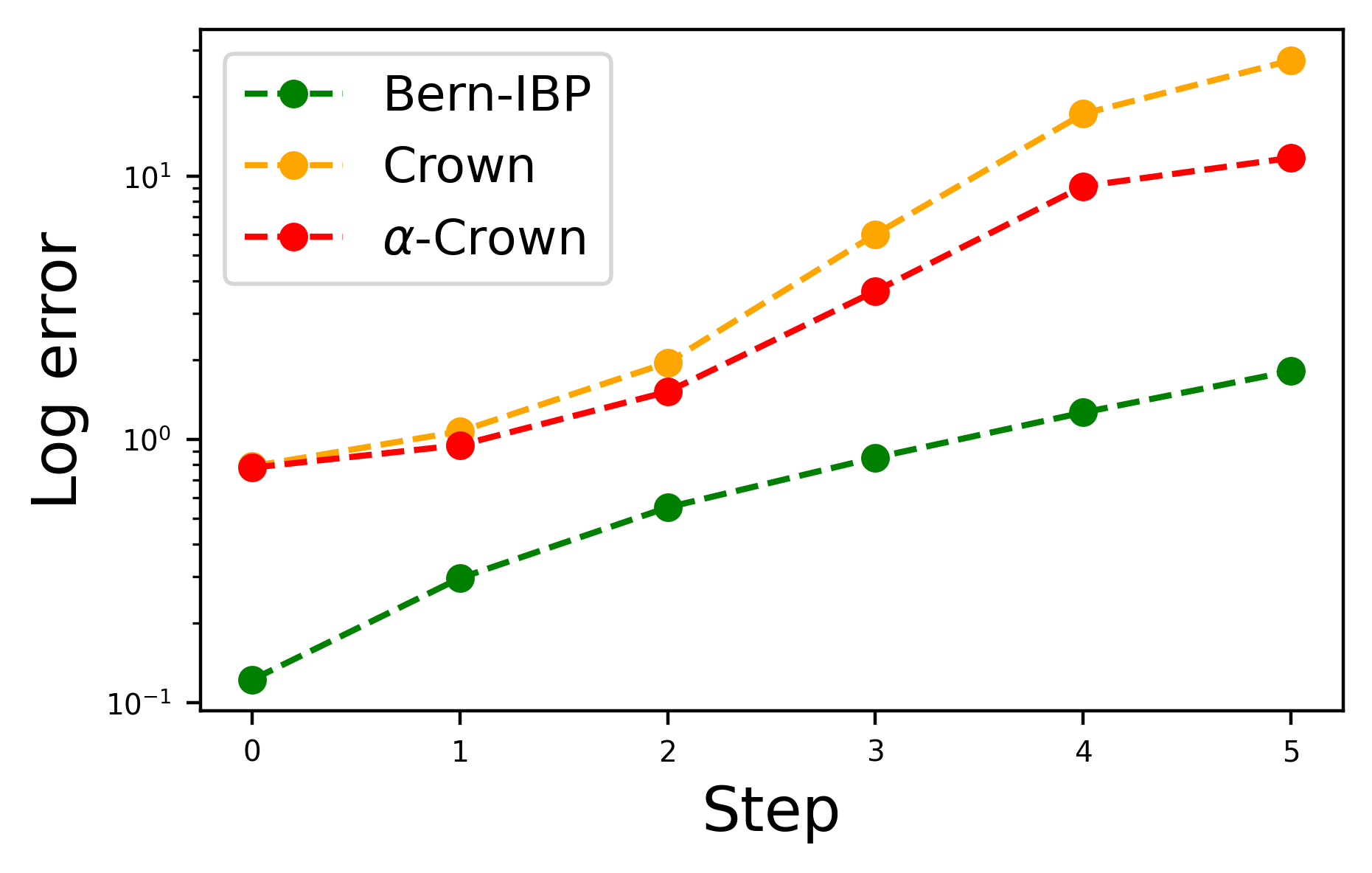}
    \end{subfigure}
    \caption{(Left) The trajectory of the Quadrotor for the ReLU and Bernstein polynomial networks. (Right) the error in the reachable set volume $e =(\hat{V} - V) / V$ for each of the networks after each step. $\hat{V}$ is the estimated volume using the respective bounding method and $V$ is the true volume of the reachable set using heavy sampling}
    \label{fig:reach_quadrotor} 
\end{figure}

After training, we perform reachability analysis with horizon $T=6$ on each network using the respective bounding methods: Crown and $\alpha$-Crown for ReLU networks and the proposed Bern-IBP for Bernstein polynomial networks. We compute the volume of the reachable sets after each step for each network. The results are visualized in Figure \ref{fig:reach_quadrotor}, comparing the error in the volume of the reachable sets for both ReLU and Bernstein polynomial networks. The error is computed with respect to the true volume of the reachable set for each network, which is computed by heavy sampling. As shown in Figure \ref{fig:reach_quadrotor}, using Bern-IBP on the NN with Bernstein polynomial can lead to much tighter reachable sets compared to SOTA bounding methods for ReLU networks. This experiment provides insights into the potential benefits of using Bernstein polynomial activations for improving the tightness of reachability bounds, which can have significant implications for neural network certification for safety-critical systems.

\section{Related work}
\paragraph{Neural Network verification.} NN verification is an active field of research that focuses on developing techniques to verify the correctness and robustness of neural networks. Various methods have been proposed for NN verification to provide rigorous guarantees on the behavior of NNs and detect potential vulnerabilities such as adversarial examples and unfairness. These methods use techniques such as abstract interpretation ~\cite{ferraricomplete}, Satisfiability Modulo Theory (SMT) ~\cite{katz2019marabou}, Reachability Analysis ~\cite{bak2021,tran2020nnv} and Mixed-Integer Linear Programming (MILP) ~\cite{lomuscio2017approach, tjeng2017evaluating, bunel2020branch, anderson2020strong}. Many tools also rely on optimization and linear relaxation techniques ~\cite{wang2021beta, khedr2021peregrinn, henriksen2021deepsplit} to speedup the verification. Another line of work ~\cite{wan2023polar, fatnassi2023bern} uses higher order relaxation such as Bernstein Polynomials to certify NNs. However, frameworks for NN verification often result in loose bounds during the relaxation process or are computationally expensive, particularly for large-scale networks.

\paragraph{Polynomial activations.} NNs with polynomial activations have been studied in ~\cite{gottemukkula2020polynomial}. Theoretical work was established on their expressiveness \cite{kileel2019expressive} and their universal approximation property \cite{kidger2020universal} is established under certain conditions. However, to the best of our knowledge, using Bernstein polynomials in Deep NNs and their impact on NN certification has not been explored yet.

\paragraph{Polynomial Neural Networks.} A recent work\cite{chrysos2021deep} proposed a new class of approximators called $\Pi$-nets, which is based on polynomial expansion. Empirical evidence has shown that $\Pi$-nets are highly expressive and capable of producing state-of-the-art results in a variety of tasks, including image, graph, and audio processing, even without the use of non-linear activation functions. When combined with activation functions, they have been demonstrated to achieve state-of-the-art performance in challenging tasks such as image generation, face verification, and 3D mesh representation learning. A framework for certifying such networks using $\alpha$-convexification was introduced in \cite{rocamorasound}.

\section{Discussion and limitations}
\paragraph{Societal impact.} The societal impact of utilizing Bernstein polynomial activations in neural networks lies in their potential to enhance the reliability and interpretability of AI systems, enabling improved safety, fairness, and transparency in various real-world applications.
\paragraph{Limitations.} While Bernstein polynomials offer advantages in the context of certification, they also pose some limitations. One limitation is the increased computational complexity during training compared to ReLU networks.
\bibliographystyle{ieeetr}
\bibliography{mybib}
\clearpage
\appendix

\section{Bernstein Polynomials}
\subsection{Proof of Proposition~\ref{prop:gradient}}
\label{app:proof}
\begin{proof}
    Before we prove our result, we review the following properties of Bernstein polynomials.

\begin{property}[Positivity ~\cite{titi2019}]
\label{prop:positivity}
Bernstein basis polynomials are non-negative on the interval $[l, u]$, i.e., $b_{n,k}^{[l,u]}(x) \geq 0$ for all $x \in [l, u]$.
\end{property}

\begin{property}[Partition of Unity ~\cite{titi2019}] 
\label{prop:unitypartition}
The sum of Bernstein basis polynomials of the same degree is equal to 1 on the interval $[l, u]$, i.e., $\sum_{k=0}^n b_{n,k}^{[l,u]}(x) = 1$, $\forall x \in [l, u]$.
\end{property}

\begin{property}[Closed under differentiation~\cite{doha2011derivatives}]
\label{prop:diff}
The derivative of an $n$-degree Bernstein polynomial is $n$ multiplied by the difference of two $(n-1)$-degree Bernstein polynomials. Concretely,
$$ \frac{d}{dx} b^{[l,u]}_{n,k}(x)= n \left(b^{[l,u]}_{n-1,k-1} (x) - b^{[l,u]}_{n-1,k}(x) \right)$$
\end{property}

Now, it follows from Property~\ref{prop:diff} that:
\begin{align*}
    \left\vert \frac{d}{dx} \sigma(x;l,u,\boldsymbol{c}) \right\vert 
    &= \left\vert \sum_{k = 0}^n c_k n \left(b^{[l,u]}_{n-1,k-1} (x) - b^{[l,u]}_{n-1,k}(x) \right) \right\vert\\
    & \le  \sum_{k = 0}^n \left \vert c_k n b^{[l,u]}_{n-1,k-1} (x)  \right\vert +  \sum_{k = 0}^n \left\vert c_k n b^{[l,u]}_{n-1,k}(x) \right\vert \\
    &\le n \max_k \vert c_k \vert \sum_{k = 0}^n \left\vert b^{[l,u]}_{n-1,k-1}(x) \right\vert +  n \max_k \vert c_k \vert \sum_{k = 0}^n \left\vert b^{[l,u]}_{n-1,k}(x) \right\vert \\
    &\stackrel{(a)}{=} n \max_k \vert c_k \vert \sum_{k = 0}^n  b^{[l,u]}_{n-1,k-1}(x) +  n \max_k \vert c_k \vert \sum_{k = 0}^n  b^{[l,u]}_{n-1,k}(x) \\
    &\stackrel{(b)}{=} n \max_k \vert c_k \vert + n \max_k \vert c_k \vert (1 + b^{[l,u]}_{n-1,n}(x))
    \stackrel{(c)}{=} 2 n \max_k \vert c_k \vert
\end{align*}
where $(a)$ follows from Property~\ref{prop:positivity}; $(b)$ follows from Property~\ref{prop:unitypartition}, and $(c)$ follows from the definition of Bernstein basis and the fact that the binomial coefficient $\binom{n-1}{n} = 0$.

Similarly, 
\begin{align*}
    \left\vert \frac{d}{dc_i} \sigma(x;l,u,\boldsymbol{c}) \right\vert 
    &= \left\vert b_{n,i}^{[l,u]}(x) \right\vert \stackrel{(d)}{=} b_{n,i}^{[l,u]}(x) \stackrel{(e)}{\le} 1.
\end{align*}
where $(d)$ follows from Property~\ref{prop:positivity} and $(e)$ follows from both Properties~\ref{prop:positivity} and~\ref{prop:unitypartition} which implies that Bernstein basis satisfy $0 \le b_{n,i}^{[l,u]}(x) \le 1$.

\end{proof}

\begin{figure}[!t]
    \centering
    \includegraphics[scale=0.5]{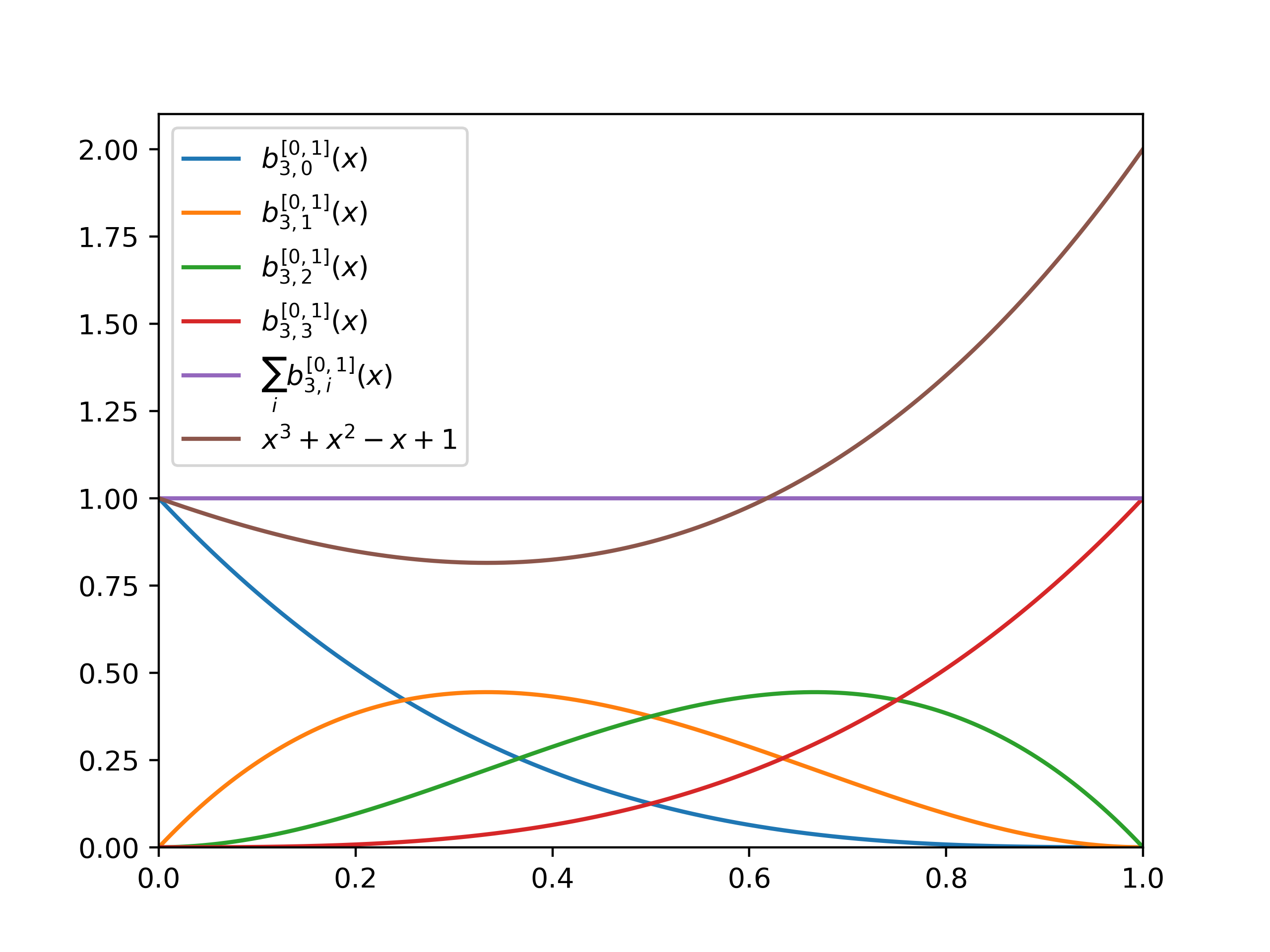}
    \caption{A visual representation of the polynomial $f(x) = x^3 + x^2 - x + 1$ along with the Bernstein basis polynomials of degree three $b_{3,i}^{[0,1]}$ for $x \in [0,1]$. The basis polynomials exhibit positivity and unity partition properties, while the range of its Bernstein coefficients bounds the range of the polynomial.}
    \label{fig:bern_props}
\end{figure}

\subsection{Example to demonstrate properties of Bernstein polynomials}

To demonstrate the properties of Bernstein polynomials, we present a simple example to represent the polynomial $f(x) = x^3 + x^2 - x + 1$ for all $x \in [0,1]$ using the Bernstein form. Any polynomial expressed in power series form can be converted to Bernstein form by employing a closed-form expression \cite{titi2019} to calculate the Bernstein coefficients. For instance, $f(x) = x^3 + x^2 - x + 1 = \sum\limits_{i=0}^{3} c_i b_{3,i}^{[0,1]}$ for $x \in [0,1]$, with $c_0=1$, $c_1=c_2=\frac{2}{3}$, and $c_3=2$. Figure \ref{fig:bern_props} illustrates a plot of the polynomial $f(x)$ and the Bernstein basis polynomials $b_{3,i}^{[0,1]}$. As depicted in the figure, the basis polynomials are positive (Property \ref{prop:positivity}) and sum to 1 (Property \ref{prop:unitypartition}). The range of the polynomial is constrained by the Bernstein coefficients' range, which is $[\frac{2}{3}, 2]$ (Property \ref{prop:enclosure}). Lastly, applying the subdivision property to compute the coefficients of the Bernstein polynomial on $[0.6,0.8]$ results in $c_0=1.352$, $c_1=1.184$, $c_2=1.0613$, and $c_3=0.976$. With the new coefficients, we can use the range enclosure property to infer that the polynomial's range on $[0.6,0.8]$ is $[0.976,1.352]$.

\section{Implementation of Bern-IBP}
\label{sec:implementation}
In this section, we discuss the implementation details of Bern-IBP and how it can be applied to certify global and local properties. 

Following the discussion in section \ref{sec:cert_glob}, we can check if a global property holds by examining the output bounds of the $\mathcal{NN}$. Algorithm \ref{alg:global_cert} provides a procedure for incomplete certification, which relies on Property \ref{prop:enclosure} to efficiently compute bounds on the output and check if the property holds. The output bounds are simply the minimum and maximum of the Bernstein coefficients of the last layer. The bounds computed using the Bernstein coefficients are not only much tighter than IBP bounds (as demonstrated in experiment \ref{exp:bounds}), but they are also more computationally efficient, as they do not require any matrix-vector operations

\begin{algorithm}[ht]
    \caption{Incomplete Certification of a \textbf{global} output property $y^{(L)} = \mathcal{NN}(\boldsymbol{y}^{(0)}) > 0$}
    \label{alg:global_cert}
\begin{algorithmic}[1]
\State Given: Neural Network $\mathcal{NN}$ with $L$ layers, and input bounds $\boldsymbol{l}^{(0)}, \boldsymbol{u}^{(0)}$
\State $l^{(L)} = \min\limits_{i} c_i^{(L)}$ 
\If{$l^{(L)} > 0$}
        \State\Return SAT
\Else
        \State\Return UNKNOWN

\EndIf
\end{algorithmic}
\end{algorithm}

Certification of local properties defined on a subset of the input domain $\mathcal{D}$ can benefit from computing tighter bounds on the outputs of the NN using Bern-IBP. Algorithm \ref{alg:local_verif} propagates the input bounds on a layer-by-layer basis, for linear and convolutional layers, we propagate the bounds using IBP. For Bernstein layers, we first apply the subdivision property to compute a new set of Bernstein coefficients to represent the polynomial on a subregion of $[\boldsymbol{l}^{(k)} ,\boldsymbol{u}^{(k)}]$, then, using the new coefficients, we apply the enclosure property to bound the output of the Bernstein activation. This procedure result in much tighter bounds compared to IBP (as shown in Experiment \ref{exp:bounds}) and Appendix \ref{app:lower_bounds}

\begin{algorithm}[ht]
    \caption{Incomplete Certification of a \textbf{local} output property $y^{(L)} = \mathcal{NN}(\boldsymbol{y}^{(0)}) > 0$}
    \label{alg:local_verif}
\begin{algorithmic}[1]
\State Given: Neural Network $\mathcal{NN}$ with $L$ layers, and input bounds $\boldsymbol{l}^{(0)}, \boldsymbol{u}^{(0)}$
\For{$i$ = $1$....$L$}
    \If{$type(layer\: i)$ is Linear or Conv}
        \State $\hat{\boldsymbol{l}}^{(i)}, \hat{\boldsymbol{u}}^{(i)} \gets \text{IBP}(layer\:  i, [\boldsymbol{l}^{(i-1)}, \boldsymbol{u}^{(i-1)}])$
    \Else
        \For{\textbf{each neuron} $z$ in $layer$ $i$} \Comment{\textcolor{red}{Actual implementation is vectorized}}
        \State $\tau \gets \frac{\hat{l}_z^{(i)} - l_z^{(i)}}{u_z^{(i)} - l_z^{(i)}}$ \Comment{\textcolor{red}{$l_z^{(i)}$ and $u_z^{(i)}$ denote lower and upper bounds for neuron $z$ for the entire input domain $\mathcal{D}$}}
            \For{$k$ = $0$....$n$}
                \For{$j$ = $k$....$n$}
                    \State $c_j^k \gets\left\{
                            \begin{array}{ll}
                                c_j  & \mbox{if } k = 0 \\
                                (1-\tau)c_{j-1}^{k-1} + \tau c_j^{k-1} & \mbox{if } k > 0
                            \end{array}
                        \right.$ 
                    \State $\boldsymbol{c'} \gets c_i^i$
                    \State $\boldsymbol{c''} \gets c_n^{n-i}$
                    \Comment{\textcolor{red}{$\boldsymbol{c''}$ are the coefficients of the polynomial on $[\hat{l}_z, ub_z]$}}
                    \LineComment{\# Lines 10 to 16 need to be executed twice to compute the coefficients of the polynomial on $[\hat{l}_z, ub_z]$. Omitted for simplicity}
                    
                \EndFor
            \EndFor
        \State $\hat{l}_z^{(i)} = \min\limits_i \boldsymbol{c''}$, $\quad \hat{u}_z^{(i)} = \max\limits_i \boldsymbol{c''}$
        \EndFor
    \EndIf
\EndFor

\If{$\hat{l}^{(L)} > 0$}
        \State\Return SAT
\Else
        \State\Return UNKNOWN

\EndIf
\end{algorithmic}
\end{algorithm}


\section{Additional information on numerical experiments}
\label{app:exp_details}
\subsection{Experimental Setup}
\label{app:exp_setup}
\paragraph{Datasets.} In our MNIST and CIFAR-10 experiments, we employ \verb|torchvision.datasets| to load the datasets, maintaining the original data splits. While we normalize the input images for CIFAR-10, we do not apply any data augmentation techniques. To evaluate the certified accuracy of our models, we utilize the test set during the certification process.

\paragraph{Certified training.} During certified training, our models are trained using the Adam optimizer \cite{kingma2014adam} for 100 epochs (unless otherwise specified) with an initial learning rate of $5\mathrm{e}{-3}$. We incorporate an exponential learning rate decay of $0.999$ that begins after 50 epochs. For the MNIST dataset, we employ a batch size of $512$, while for CIFAR-10, we use a batch size of 256, except for larger models where a batch size of $128$ is utilized. Prior to incorporating the robust loss into the objective, we perform 10 warmup epochs for MNIST and 20 for CIFAR-10. The total loss comprises a weighted combination of the natural cross-entropy loss and the robust loss. The weight follows a linear schedule after the warmup phase, gradually increasing to optimize more for the robust loss towards the end of training. In terms of evaluation, the primary metric is certified accuracy, which represents the percentage of test examples for which the model can confidently make correct predictions within the given $l_\infty$ perturbation radius.

\paragraph{Bernstein activations.} We use the same value of the hyperparameter $n$ for all neurons in the network. For a Bersntein activation layer with $m$ neurons, we initialize the Bernstein coefficients from a normal distribution $c_k \sim \mathcal{N}(\boldsymbol{0},\sigma^{2})$, where $\sigma^2 = \frac{1}{m}$.

\subsection{Models Architecture}
\label{app:arch}
Table \ref{tab:arch} lists the architecture, polynomial order and number of parameters for the Neural networks used to compare the certified robustness with ReLU networks from SOK\cite{li2020sok} benchmark.
\begin{table}[H]
\centering
\caption{Neural Network Models}
\label{tab:arch}
\resizebox{\textwidth}{!}{%
\begin{tabular}{c|ccc|ll}
\multirow{2}{*}{Model} & \multirow{2}{*}{Structure} & \multicolumn{2}{c|}{Degree} & \multicolumn{2}{c}{\# of Parameters} \\
      &                                              & MNIST & CIFAR-10 & \multicolumn{1}{c}{MNIST} & \multicolumn{1}{c}{CIFAR-10} \\ \hline
FCNNa & {[}20,20,10{]}                               & 4     & 3        & 16,530                    & 62,250                       \\
FCNNb & {[}100,100,100,10{]}                         & 8     & 3        & 102,410                   & 329,710                      \\
FCNNc & {[}100,100,100,100,100,100,100,10{]}         & 10    & 10       & 147,810                   & 376,610                      \\
CNNa  & {[}CONV16,CONV16,100,10{]}                   & 10    & 12       & 219,250                   & 296,090                      \\
CNNb  & {[}CONV16,CONV16,CONV32,CONV32,512,10{]}     & 4     & 8        & 953,946                   & 1,360,922                    \\
CNNc  & {[}CONV32,CONV32,CONV64,CONV64,512,512,10{]} & 2     & 7        & 2,118,954                 & 2,966,570                   
\end{tabular}%
}
\end{table}

\subsection{Training time}
\label{app:timing}
In this section, we study the computational complexity of training DeepBern-Nets.

\begin{figure}[ht]
    \centering
    \includegraphics[width=0.45\textwidth]{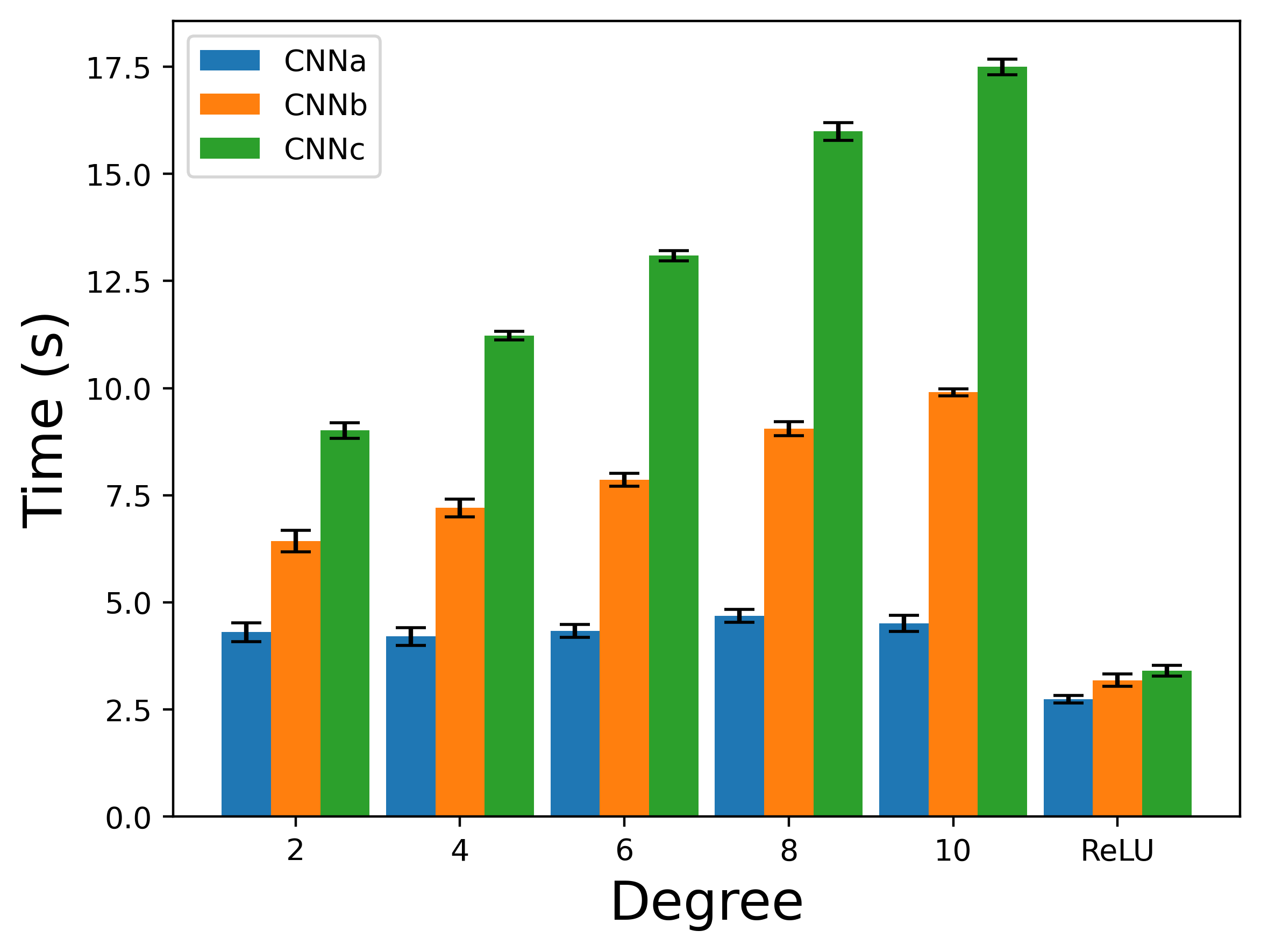}
    \includegraphics[width=0.45\textwidth]{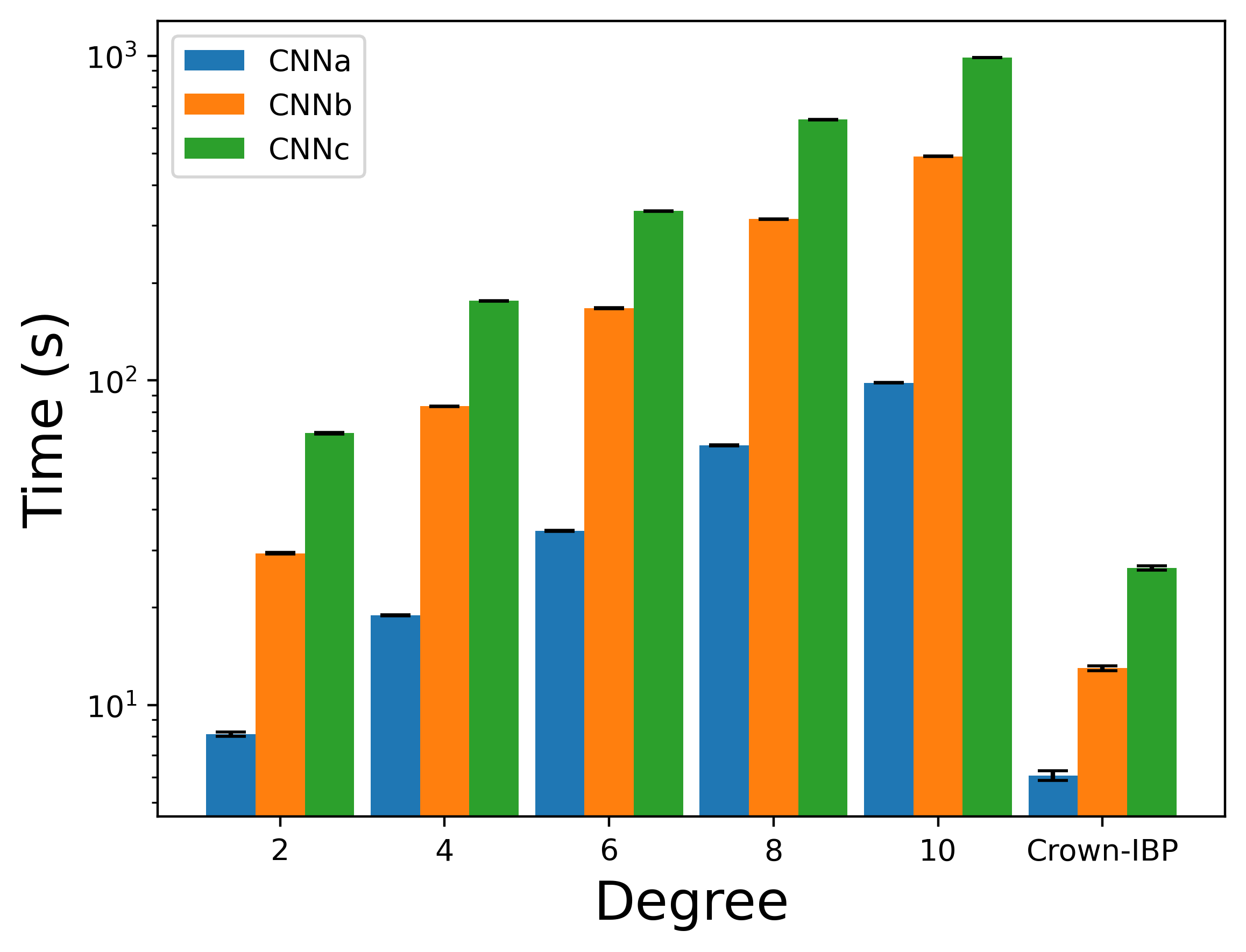}
    \caption{(Left) Training time (per epoch) for three different model architectures and increasing order of Bernstein polynomials. (Right) training time (per epoch) for training networks with certified training objective functions (i.e., Bern-IBP must be used with every epoch to compute the loss in the certified training loss) and increasing Bernstein order on the MNIST dataset. Crown-IBP execution times are reported for ReLU networks with the same architecture.}
    \label{fig:exec_time}
\end{figure}

Figure \ref{fig:exec_time} (left) shows the average epoch time  and the standard deviation for training DeepBern-Nets. We trained NNs with three different architectures and with increasing Bernstein activation order on the MNIST dataset. The figure shows that for each architecture, the training time seems to grow linearly with the polynomial order (used in the activation functions), except for the small architecture (CNNa). This is due to the fact that higher-order polynomials introduce more parameters into the network and the fact that the cost of computing the Bernstein bounds during training also scales with the order of the polynomial. We also report the training time of a ReLU network with the same architecture to contrast an important underlying trade-off; Bernstein activations are trained with certifiability in mind, which comes with the extra computational cost during training.

Figure \ref{fig:exec_time} (right) shows the average epoch time ad the standard deviation for certified training of DeepBern-Nets using Bern-IBP. We also report the certified training epoch time for ReLU networks of the same architecture using Crown-IBP. We observe a similar trend of linear increase in training time with increasing the order of Bernstein activation.


\subsection{Tightness of output bounds - Bern-IBP vs IBP}
\label{app:lower_bounds}
In this section, we complement Experiment \ref{exp:bounds} by reporting the raw data for computing the lower bound on the robustness margin as defined in \ref{eq:robustness_lb} computed using Bern-IBP and IBP on the MNIST dataset. Tables \ref{tab:lower_bounds_raw}, \ref{tab:median_lower_bounds_raw}, \ref{tab:min_lower_bounds_raw}, and \ref{tab:max_lower_bounds_raw} present the mean, median, minimum, and maximum values for the lower bounds using both methods on NNs of increasing order and different values of $\epsilon$, respectively. The model architecture is CNNb as described in \ref{app:arch}. The tables clearly demonstrate that Bern-IBP achieves significantly higher precision than IBP in bounding DeepBern-Nets. This improvement is observed consistently across all DeepBern-Nets orders and various epsilon values. Bern-IBP outperforms IBP by orders of magnitude, highlighting its effectiveness in providing tighter bounds.
\begin{table}[ht]
\centering
\caption{
Raw values of the average of $\mathbb{L}_\text{robust}$ in Experiment 1.1. The results show that Bern-IBP results in  orders of magnitude tighter values for $\mathbb{L}_\text{robust}$ compared with IBP.
}
\label{tab:lower_bounds_raw}
\resizebox{\textwidth}{!}{%
\begin{tabular}{c|cc|cc|cc|cc}
\toprule
\multirow{2}{*}{Order} &
  \multicolumn{2}{c|}{$\epsilon=0.001$} &
  \multicolumn{2}{c|}{$\epsilon=0.01$} &
  \multicolumn{2}{c|}{$\epsilon=0.04$} &
  \multicolumn{2}{c}{$\epsilon=0.1$} \\
  & IBP          & Bern-IBP    & IBP          & Bern-IBP      & IBP           & Bern-IBP      & IBP         & Bern-IBP     \\ \hline
2 & 3.25  & 6.44 & -23.75 & 0.33  & -47.85  & -4.39  & -48.10 & 0.02 \\
3 & -31.35  & 6.91 & -145.52 & -0.30 & -13175.76  & -8.39  & -2.16e+8  & -104.32 \\
4 & -109.46 & 6.75 & -1779.87 & -0.33 & -8.4e+11 & -0.38 & -2.53e+21   & -8.36 \\
5 & -410.41 & 6.94 & -2.65e+31    & 2.67   & -inf           & -0.40 & -inf         & -7.69 \\
6 & -2429.93 & 7.05 & -inf          & 1.13   & -inf           & -11.63  & -inf         & -42.75 \\
\bottomrule
\end{tabular}%
}
\end{table}
\begin{table}[ht]
\centering
\caption{
Raw values of the median of $\mathbb{L}_\text{robust}$ in Experiment 1.1. The results show that Bern-IBP results in  orders of magnitude tighter values for $\mathbb{L}_\text{robust}$ compared with IBP.
}
\label{tab:median_lower_bounds_raw}
\resizebox{\textwidth}{!}{%
\begin{tabular}{c|cc|cc|cc|cc}
\toprule
\multirow{2}{*}{Order} &
  \multicolumn{2}{c|}{$\epsilon=0.001$} &
  \multicolumn{2}{c|}{$\epsilon=0.01$} &
  \multicolumn{2}{c|}{$\epsilon=0.04$} &
  \multicolumn{2}{c}{$\epsilon=0.1$} \\
  & IBP      & Bern-IBP & IBP       & Bern-IBP & IBP       & Bern-IBP & IBP       & Bern-IBP \\ \hline
2 & 3.38     & 6.54     & -23.6     & 0.43     & -46.71    & -4.14    & -47.25    & 0.19     \\
3 & -31.01   & 7.07     & -145.36   & -0.03    & -12108.94 & -8.14    & -1.99e+8  & -104.78  \\
4 & -105.52  & 6.9      & -1584.41  & -0.1     & -2.12e+11 & -0.13    & -1.91e+20 & -8.36    \\
5 & -404.44  & 7.17     & -1.88e+10 & 2.96     & -2.34e+34 & -0.17    & -inf      & -7.93    \\
6 & -2334.94 & 7.22     & -1.32e+24 & 1.56     & -inf      & -11.25   & -inf      & -42.37  \\
\bottomrule
\end{tabular}%
}
\end{table}
\begin{table}[ht]
\centering
\caption{Raw values of the minimum of $\mathbb{L}_\text{robust}$ in Experiment 1.1. The results show that Bern-IBP results in  orders of magnitude tighter values for $\mathbb{L}_\text{robust}$ compared with IBP.}
\label{tab:min_lower_bounds_raw}
\resizebox{\textwidth}{!}{%
\begin{tabular}{c|cc|cc|cc|cc}
\toprule
\multirow{2}{*}{Order} &
  \multicolumn{2}{c|}{$\epsilon=0.001$} &
  \multicolumn{2}{c|}{$\epsilon=0.01$} &
  \multicolumn{2}{c|}{$\epsilon=0.04$} &
  \multicolumn{2}{c}{$\epsilon=0.1$} \\
  & IBP        & Bern-IBP & IBP         & Bern-IBP & IBP          & Bern-IBP & IBP          & Bern-IBP \\ \hline
2 & -20.16     & -16.63   & -42.72      & -16.56   & -83.7        & -22.22   & -71.33       & -8.25    \\
3 & -96.55     & -12.16   & -205.09     & -14.02   & -34962.84    & -22.91   & -2302369792  & -137.07  \\
4 & -3550.07   & -10.15   & -56758.56   & -13.72   & -1.09065E+15 & -9.23    & -8.24695E+24 & -23.03   \\
5 & -1345.89   & -11.78   & -2.2861E+35 & -12.93   & -inf         & -8.68    & -inf         & -18.11   \\
6 & -109130.05 & -12.24   & -inf        & -17.03   & -inf         & -30.47   & -inf         & -72.53  \\
\bottomrule
\end{tabular}%
}
\end{table}
\begin{table}[!h]
\centering
\caption{
Raw values of the maximum of $\mathbb{L}_\text{robust}$ in Experiment 1.1. The results show that Bern-IBP results in  orders of magnitude tighter values for $\mathbb{L}_\text{robust}$ compared with IBP.
}
\label{tab:max_lower_bounds_raw}
\resizebox{\textwidth}{!}{%
\begin{tabular}{c|cc|cc|cc|cc}
\toprule
\multirow{2}{*}{Order} &
  \multicolumn{2}{c|}{$\epsilon=0.001$} &
  \multicolumn{2}{c|}{$\epsilon=0.01$} &
  \multicolumn{2}{c|}{$\epsilon=0.04$} &
  \multicolumn{2}{c}{$\epsilon=0.1$} \\
  & IBP     & Bern-IBP & IBP          & Bern-IBP & IBP          & Bern-IBP & IBP          & Bern-IBP \\ \hline
2 & 14.83   & 18.01    & -10.33       & 11.24    & -23.36       & 5.61     & -28          & 5.42     \\
3 & -12.53  & 20.98    & -96.18       & 7.18     & -2952.79     & 1.21     & -13615902    & -76.27   \\
4 & -71.81  & 18.97    & -767.27      & 7.59     & -307653536   & 4.35     & -1.80601E+17 & 1.4      \\
5 & -249.03 & 16.76    & -8781314     & 12.62    & -7.72777E+25 & 4.26     & -inf         & 2.62     \\
6 & -1055.1 & 19.99    & -1.13219E+15 & 10.94    & -inf         & 0.18     & -inf         & -19.47  \\
\bottomrule
\end{tabular}%
}
\end{table}

\end{document}